\documentclass{article}

\usepackage{arxiv}

\usepackage[utf8]{inputenc} 
\usepackage[T1]{fontenc}    
\usepackage{hyperref}       
\usepackage{url}            
\usepackage{booktabs}       
\usepackage{amsfonts}       
\usepackage{nicefrac}       
\usepackage{microtype}      
\usepackage{lipsum}
\usepackage{graphicx}
\graphicspath{ {./images/} }
\usepackage{siunitx}
\usepackage{booktabs}
\usepackage[utf8]{inputenc}
\usepackage[labelfont=bf]{caption}
\usepackage{threeparttable}
\usepackage{bm}
\usepackage{physics}
\usepackage{gensymb}
\usepackage{fancyhdr}
\usepackage{subcaption}
\usepackage{balance}
\usepackage{amsmath}
\usepackage{makecell}

\title{Motion Planning and Tracking Control of Unmanned Underwater Vehicles: Technologies, Challenges and Prospects}

\author{
 Danjie Zhu, Tao Yan and Simon X. Yang \\
  School of Engineering, University of Guelph\\ 
  Guelph, ON, Canada, N1G2W1\\
  \texttt{\{danjie; tyan03; syang\}@uoguelph.ca} \\
}

\begin{document}
\maketitle
\begin{abstract}
The motion planning and tracking control techniques of unmanned underwater vehicles (UUV) are fundamentally significant for efficient and robust UUV navigation, which is crucial for underwater rescue, facility maintenance, marine resource exploration, aquatic recreation, etc. Studies on UUV motion planning and tracking control have been growing rapidly worldwide, which are usually sorted into the following topics: task assignment of the multi-UUV system, UUV path planning and UUV trajectory tracking. This paper provides a comprehensive review of conventional and intelligent technologies for motion planning and tracking control of UUVs. Analysis of the benefits and drawbacks of these various methodologies in literature is presented. In addition, the challenges and prospects of UUV motion planning and tracking control are provided as possible developments for future research.
\end{abstract}

\section{Introduction}
More attention has been concentrated on underwater navigation since this century due to the abundant resources buried in the deep-sea area, such as biological, mineral and space resources \cite{r4}. Therefore, underwater vehicles (UV) have been applied due to their adaptiveness and safety when exploring undersea environments. The vehicle can tackle the problems of hardly-predictable obstacles, currents flow and hydraulic pressure as well as provide longer operating time and more functions than the human divers \cite{r5,r6}. 

Meanwhile, the unmanned underwater vehicle (UUV) is developed, which can be divided into the remote-operated vehicle (ROV) or the autonomous underwater vehicle (AUV). The motion planning and tracking control of UUVs are assumed to be significant technologies for accomplishing efficient underwater navigation with a guaranteed response time and without direct contact with marine dangers. Research on the underwater motion planning and tracking control of UUVs originated decades ago and has been under the spotlight in recent years. The history of the UUV can be traced back to the mid-last century when an unmanned vehicle was invented by the US Navy for recovering the hydrogen bomb lost on the coast of Spain \cite{USnavy}. In 2009, the success of finding crack pieces of Air France 447 realized by the UUV has verified the vehicle's promising application in underwater navigation. In 2014, the search of the flight in the MH370 accident also brought out the growing attention on the demanding underwater navigation, which highly depends on the technology of the UUV motion planning and tracking control \cite{MH370}. Nowadays, detecting deep into the ocean area for digging more available resources such as undersea oil development also requires the continuous progress of the UUV motion planning and tracking control \cite{advanceofuuv}.. 

According to the statistics collected by Web of Science, the number of organizations that devote efforts to underwater vehicle research has impressively increased in the past years. This trend corresponds to the growing demand for underwater navigation worldwide, where studies of underwater motion planning and tracking control are regarded as hot topics among UUV research projects. The underwater motion planning and tracking control form the most crucial part of underwater navigation. The UUV motion planning is established on conventional or intelligent technologies of vehicle posture planning, task assignment and path planning; while the UUV tracking control is mainly about vehicle trajectory tracking. The paper focuses on UUV task assignment, path planning and trajectory tracking controls. The task assignment is designed for the multi-UUV system, where multiple vehicles shall be arranged simultaneously to achieve the most efficient collective navigating plan without mutual interference; the path planning of the UUV aims at giving the optimal instruction to the vehicle for arriving at the target, which can largely save the time and reduce the energy consumption; the trajectory tracking study of the UUV guarantees the robustness and manner of the vehicle operation in practical cases. However, studies related to the underwater motion planning and tracking control have not been thoroughly investigated due to the complexity of the ocean environment and the vehicle system \cite{2018Sun, huangreview}. In addition, to overcome the difficulty of accomplishing complex underwater operations, the multi-UUV system, which refers to the system of multiple UUVs and multiple targets, has pulled great attention owing to its high parallelism, robustness and efficiency \cite{petillo2014,Panda2020,Hadi2021}. 

Motivated by the goal of realizing efficient and robust UUV navigation in ocean environments, studies related to the UUV motion planning and tracking controls should be systematically surveyed and discussed to address their potential in applications. Meanwhile, the progress in this field can be promoted by analyzing the deficiency and possible developments of the relevant technologies. Therefore, the contribution of the paper is to collect and analyze technologies that have been and can be applied on the motion planning and tracking control of UUVs. The benefits and drawbacks of these technologies have been concluded and challenges and prospects are derived based on the gap in the literature. These analyses and conclusions are supposed to provide a brief overview of studies that can be developed on certain issues for researchers at the entry level in the field of UUV motion planning and tracking control.

In this paper, a brief review of the technologies regarding the UUV motion planning and tracking control is proposed. The review investigates the current development of the motion planning and tracking control achieved by the UUV and then derives the challenges as well as possible prospects of the study. The introduction is given in the first section. In the second and third sections, the current research status is described. Methodologies of UUV motion planning are organized into divisions of task assignment and path planning; meanwhile, in the tracking control section, the trajectory tracking methods of the UUV are surveyed. In the fourth section, challenges and possible prospects are concluded and discussed.

\section{Technologies of UUV Motion Planning}
In this section, technologies on the motion planning of UUVs are presented. Motion planning of UUVs can be mainly categorized into steps of task assignment and path planning, where the path planning usually splits into point-to-point path planning and full coverage path planning. 

\begin{figure}[htb]
  \centering
  \includegraphics[width=0.65\textwidth]{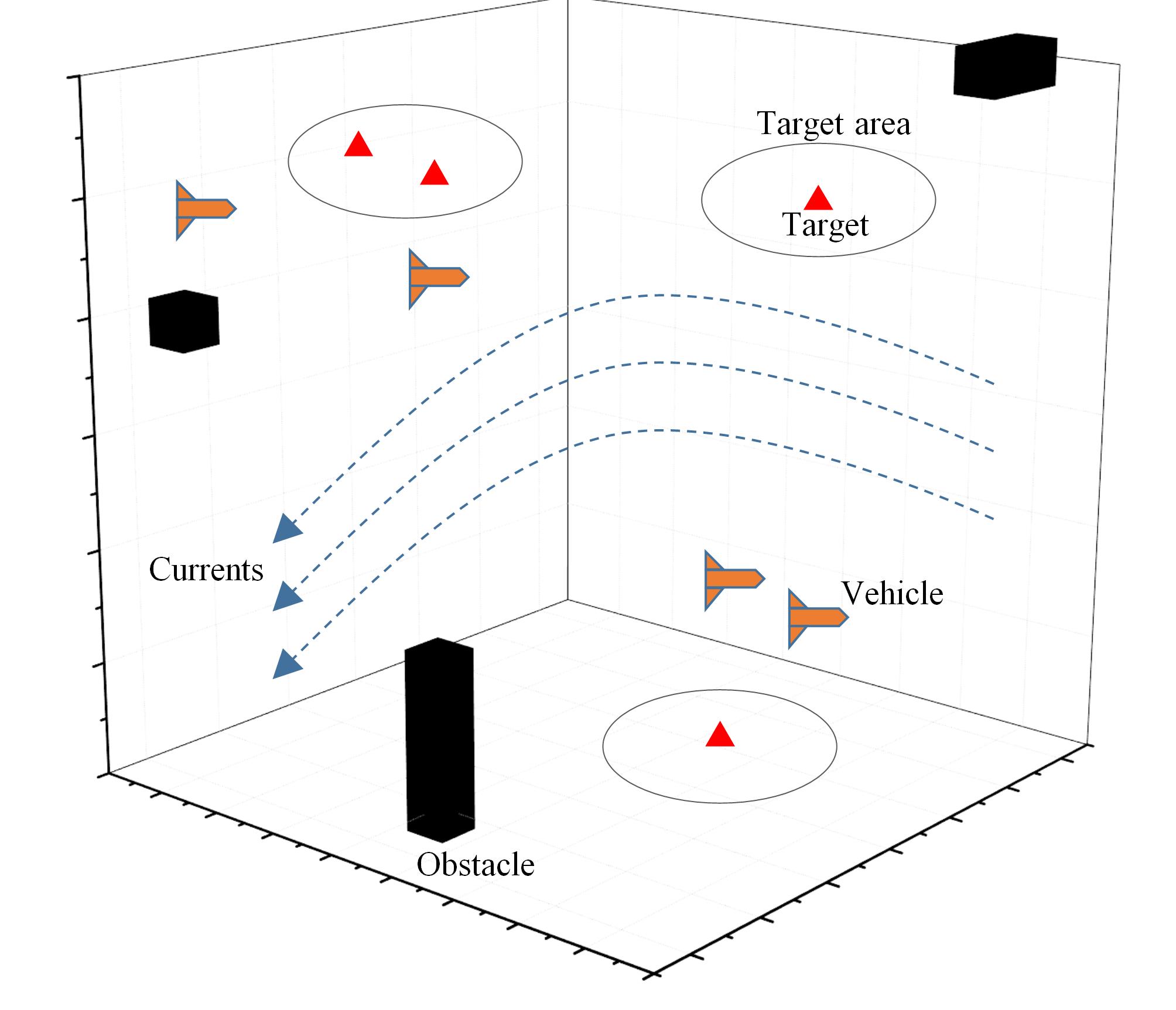}
  \caption{The underwater motion planning scenario of the UUV.}
  \label{fig:fig1}
\end{figure}

Underwater motion planning is the crucial part that decides the efficiency of a UUV navigation. The optimal vehicle motion has to be addressed in the requirement of the shortest total distance and time to arrive at the target. As the underwater motion planning scenario shown in \autoref{fig:fig1}, under the effect of ocean currents and obstacles, for the multi-UUV system, the optimal task assignment between multiple vehicles (in orange) upon multiple targets (in the red triangle) is considered as the preparation for assessing satisfactory planned paths. For the UUV path planning, the point-to-point path planning decides the initial navigation path from the vehicle to the target, while the full-coverage path planning instructs the vehicle's traversing operation after arriving at the target area (area within the black circle).

\subsection{Task Assignment of Multi-UUV System}
Originated from the last century, strategies usually applied to the task assignment of the multi-UUV system are mostly realized by directly imitating the animal behaviors in early times. These assignments are designed through sensor-collected information and the vehicle tasks are arranged referring to actual creature grouping behaviors \cite{ParkerPhdThesis,Kulkarni2010}. Mataric and his colleagues proposed a task assignment algorithm that imitates the animal grouping behaviors such as swarming and distributing \cite{Mataric1992}. Parkers established a distributed system that divided the assignment into smaller computing sections based on vehicle behaviors \cite{ParkerPhdThesis}. Miyata developed a behavior-based algorithm that independently assigned the task for vehicles based on the time priority \cite{Miyata2002}. These studies verify the directness, simple operating procedure and no delays of behavior-based algorithms. However, they stay at the at low administrative levels of imitation, which are short of self-regulation/optimization and the unsatisfactory collaboration leads to the inefficiency of the algorithm and the requirement of intelligent task assignment methods.

Agent-based algorithms have been commonly applied to the task assignment of the multi-vehicle system \cite{Turner1998,Dia2002,Ahmed2005}. In the agent-based task assignment algorithm, the whole system is assumed to be an economy entity while each vehicle works as an agent. The agent-based algorithms are regarded as decentralized approaches, as each vehicle agent is supposed to know its requirement and limitation, and the final solution is deducted based on the balance between them. The task assigned to each agent is balanced after the repeated computation and comparison of the cost to their targets, therefore, the minor consumption and the largest profit for the whole entity can be obtained at the end \cite{Akkiraju2001}. The agent-based algorithms such as the auction algorithm resolve the task assignment problem of known targets efficiently, however, they do not work well in the vehicle assignment problem of unknown targets \cite{Atkinson2005,Whal2016}. Yao applied the biased min-consensus (BMC) method, which introduces the edge weight into the standard min-consensus protocol. Yao achieved the path planning of simultaneous arrival for all UUVs through this agent-based task assignment algorithm, yet situation of unknown targets is still not developed \cite{Yao2019,Yao2020}.

\begin{figure}[!t]
  \centering
  \includegraphics[width=0.4\textwidth]{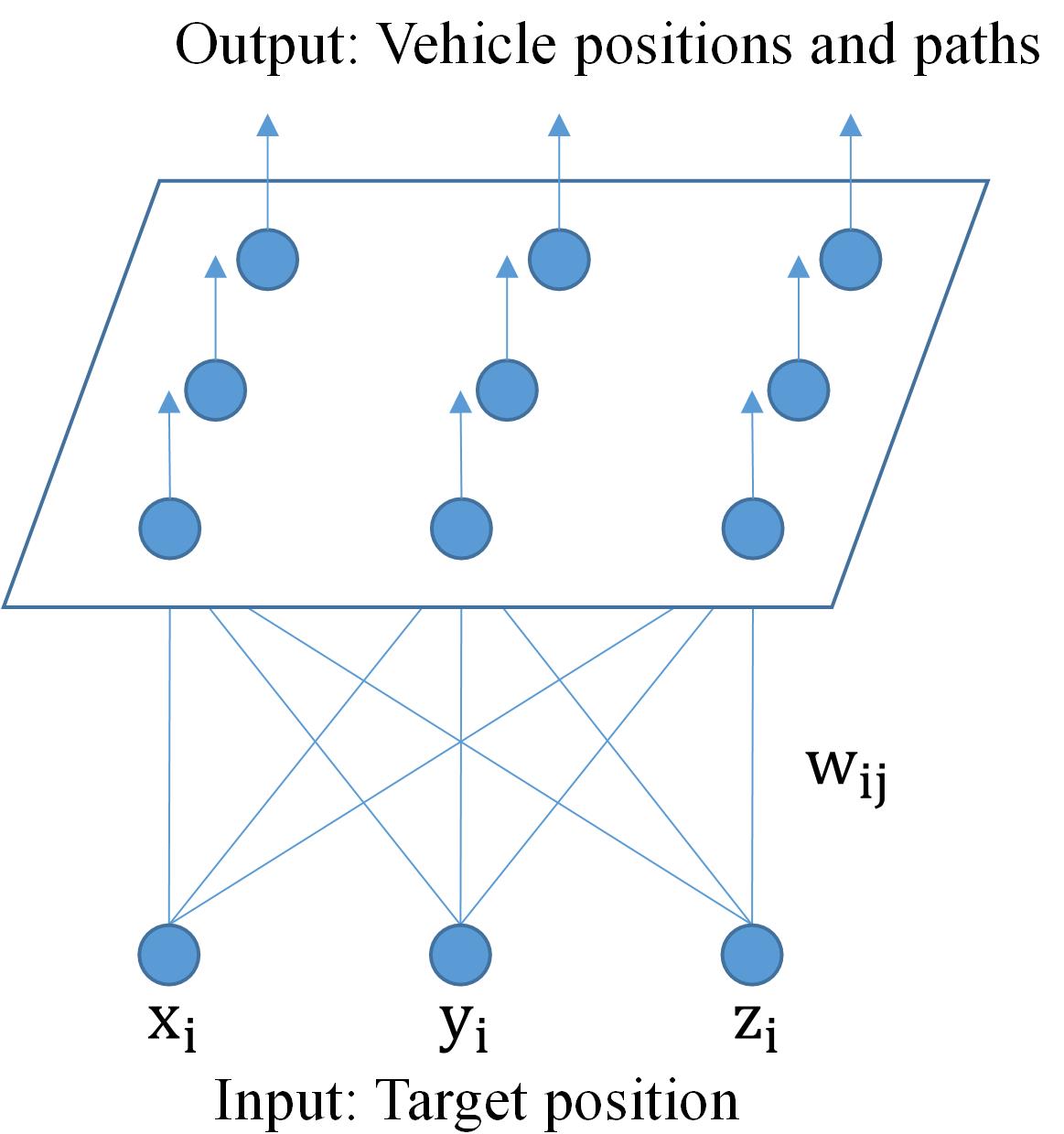}
  \caption{Structure of the SOM algorithm.}
  \label{fig:fig2}
\end{figure}

Intelligent methods such as swarm intelligence, genetic algorithm (GA), neural network (NN) have been tested in solving the problem of the multi-vehicle task assignment \cite{Tolmidis2013,Boveiri2017,Chun2015}. These intelligent methods find the best task assignment solution by the objective function established on the total searching length and the heuristic cost through iteration algorithms. In recent years, the self-organizing map (SOM), an NN-based algorithm, was applied to the task assignment problem of a multi-vehicle system due to the competitiveness and self-improving features of the neural network \cite{Kohonen1982}. The SOM-based task assignment algorithm guarantees that each vehicle in the multi-vehicle system can navigate along the shortest path to their target while maintaining the shortest total navigation cost for the whole system, whose structure is shown in \autoref{fig:fig2}. The target locations serve as the inputs while the vehicle positions and paths are the outputs of the network and the networks is updated by the weights between layers that deducted based on distances between targets and vehicles \cite{Zhu2006}. The turning direction angle and turning radius of the vehicle are then involved on the basis of the SOM method due to the vehicle's practical requirement of reducing the energy cost by reaching the target in a smooth curve in the task assignment problem \cite{Zhu2010,Huang2014}. 

However, the task assignment algorithms considering the underwater environment are still not thoroughly investigated due to the complex environmental factors and the nonlinear UUV system. Considering the complexity of the underwater environment such as the currents effect, Chow proposed an improved K-means algorithm to simultaneously resolve the task assignment and path planning problems for the multi-UUV system under the static ocean currents effect, where the vehicle successfully reached the target along smooth curves on the basis of optimal task assignment \cite{ChowPhdThesis}. Nevertheless, the method does not work well for moving targets and it lacks the discussion of applications under 3D static ocean currents effect as well as the dynamic currents condition. Zhu and his colleagues introduced SOM into the multi-UUV system and combined SOM with a velocity synthesis algorithm, hence the task assignment and path planning problem for the  multi-UUV system under time-varying ocean currents when chasing both static or dynamic targets can be addressed, which resolved the issues that existed in Chow's study, yet both the SOM-based methods cannot realize satisfactory collision avoidance\cite{Zhu2013}.

Methods that have been applied on the task assignment of the multi-UUV system are listed in Table 1, details of various intelligent methods for task assignment of UUVs can be found in Section 2.1. Gaps are still left for relative studies, which can be mainly concluded into two problems. The first problem is the difference among heterogeneous UUVs. They have different model parameters, navigating velocities or safe distances such that the assignment of parameters for every single UUV is not consistent in the practical application. The other problem is the effect of the underwater environmental factors such as the obstacles and the fluid effect, which may produce inevitable deviations or overmuch dynamic requirements for vehicles in the task assignment.  
\begin{table}[!t]
  \centering
  \footnotesize
  \caption{Algorithms for task assignment of multi-vehicle system}
  \label{tab:table1}
  \begin{tabular}{>{\raggedright}p{1.8cm}p{6.5cm}p{3.25cm}p{3.2cm}}
    \toprule
    $\textbf{Algorithms}$ & $\textbf{Logic}$ & $\textbf{Benefits}$ & $\textbf{Drawbacks}$\\
    \midrule 
Behavior Imitation Algorithms [12-15] & \makecell[l]{Simple imitation of the animal (including human)\\ grouping behaviors such as swarming and distribut-\\ing behaviors.}& \makecell[l]{1) Easy to implement; \\2) React without lags.} & \makecell[l]{1) Low efficiency; \\2) Cannot regulate them-\\selves; \\3) Difficult to optimize.}\\ \addlinespace
Agent-based Algorithms [16-23] & \makecell[l]{1) Assume the whole system as an economy entity \\while each vehicle works as an agent; \\2) Assign the task to each agent in the goal of gain-\\ing lowest cost for the whole entity.} & \makecell[l]{1) Easy to implement; \\2) Satisfactory efficiency \\when resolving problems \\of known targets.} & \makecell[l]{Do not work well in the \\task assignment of un-\\known targets.}\\ \addlinespace
Intelligent Algorithms (GA, NN-based) [24-30] & \makecell[l]{1) Regard the task assignment as a search optimiza-\\tion problem; \\2) Take the searching distance as the objective func-\\tion; \\3) Optimize through iterations.} & \makecell[l]{Outstanding adaptiveness \\due to consideration of the \\UUV system or environ-\\mental factors in the objec-\\tive function.} & \makecell[l]{1) Unsatisfactory real-time \\reaction owing to the com-\\putation complexity; \\2) Local minimum.}\\
    \bottomrule
  \end{tabular}
\end{table}

\subsection{Path Planning of UUV}
In this section, current methodologies developed for the path planning of the UUV system under different application cases are presented and concluded, in sections of point-to-point path planning and full-coverage path planning. 

\subsubsection{Point-to-point Path Planning}
After completion of the task assignment, the UUV is required to navigate to the supposed destination position from its current position with: 1) an optimized path of shortest distance, 2) avoidance of obstacles, which is described as the point-to-point path planning problem. Conventional map building methods such as grid-based modeling and topological approaches are used in the point-to-point path planning, and nowadays, typical methods that are applied in the UUV point-to-point path planning also include artificial potential field method and a wide range of intelligent path planning algorithms.

\paragraph{Map building Method}
Map building methods plan the path by mapping the vehicle's surrounding area and then deriving the optimal solution accordingly. Based on the area information collected by the vehicle sensors such as the obstacle occupied status, different methodologies of mapping these areas can be addressed and deduct an efficient path solution accordingly. The fundamental part of map building methods such as mapping the vehicle searching area usually serves as the basis of most path planning algorithms, such as intelligent algorithms, yet in this section the map building methods limit to those directly deduct on the map form without the combination of complex strategies like self-regulation or self-evolution.

\begin{figure}[htb]
  \centering
  \includegraphics[width=0.8\textwidth]{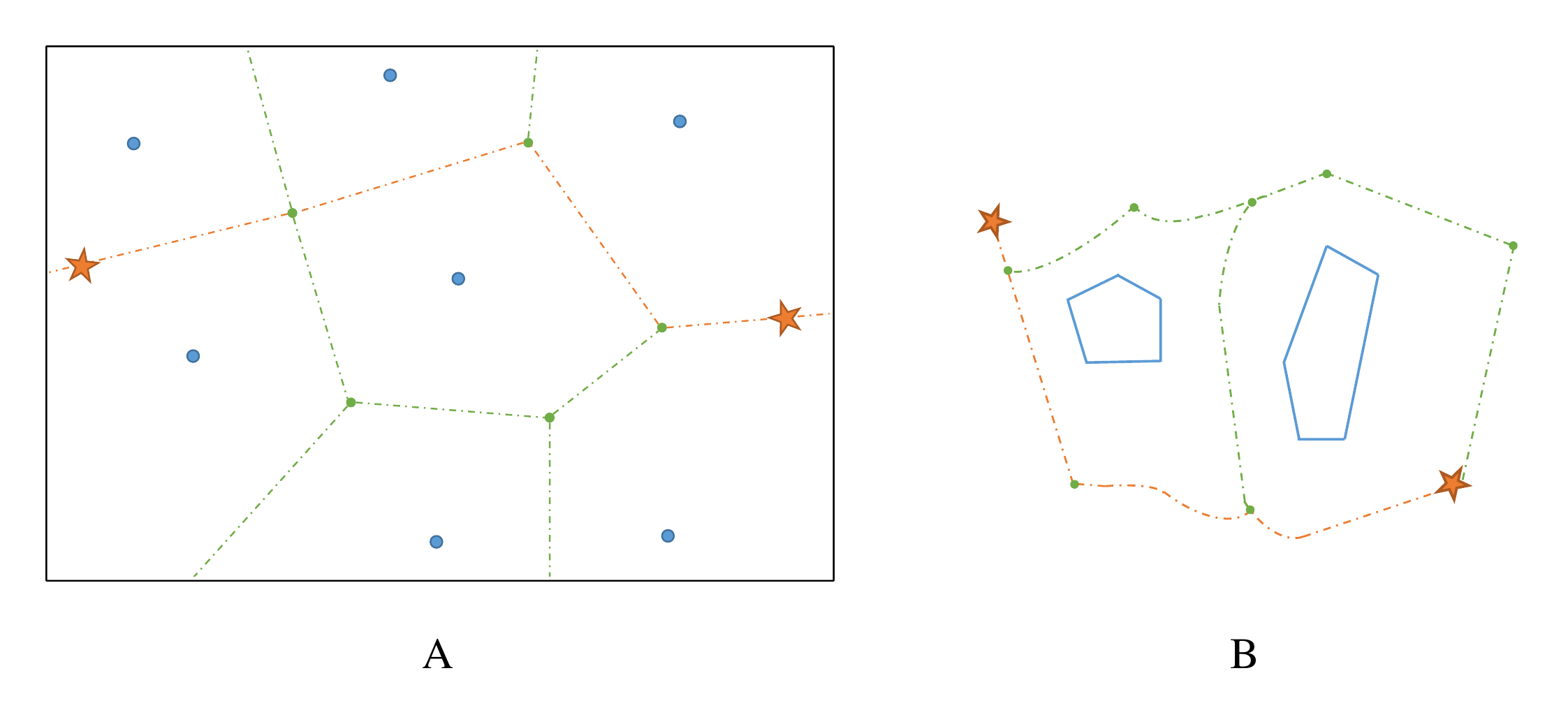}
  \caption{Typical Voronoi diagrams with the indicated graph branch nodes and optimal point-connected path, orange stars: starts and ends of the path. A: Obstacles in points; B: Obstacles in convex polygons. Orange stars: Starts and ends of the path.}
  \label{fig:fig2.5}
\end{figure}

One typical map building method is called visibility graph approach, where the graph is established on the connection of the vehicle, polygonal obstacle vertex and the destination without crossing the obstacles \cite{DAmato2021}. The optimal path is determined by finding the route between the origin point and the destination point that has the shortest distance. The visibility graph approach derives the shortest path yet it consumes long searching time and lacks flexibility, as the graph has to be reconstructed once the environmental information changes, such as the destination position or the obstacle shapes. Moreover, the visibility graph approach does not work for circular obstacles. The tangent graph method gives a more efficient path planning solution of shorter distance by controlling the vehicle to navigate along the tangent lines of obstacles \cite{Lam2005}. However, the vehicle needs to approach the obstacle as close as possible when navigating along the tangent lines such that collisions might be produced in practical cases. The Voronoi diagram method resolves the collision problem through the combination of lines and parabolas, as shown in \autoref{fig:fig2.5} where the line is defined by the vertex of obstacles while the parabola is defined by a vertex and a sideline of obstacle \cite{Magid2020,Jiankun2020}. 

Grid-based path planning methods are also a kind of widely used map building method. It decomposes the surrounding area into nonoverlapping but connected cells and then the optimal path is addressed between the origin and the destination cells without collisions. Dijkstra algorithm is one of the earliest grid-based path planning methods where a global search on all possible path solutions are required such that large computation is inevitable \cite{Dijkstra}. Therefore, A* algorithm is proposed with the advancement of adding the heuristic cost to reduce the searching space \cite{Astar1}. However, typical underwater disturbances such as the effect of currents might bring inevitable influence for UUV path planning, hence the traditional grid-based path planning methods that need map of high accuracy and consistency are not appropriate to the UUV system \cite{RN1345,RN1338}.

\paragraph{Artificial Potential Field Method}
The artificial potential field (APF) method is established on a virtual artificial potential field predefined. The proposed destination is determined as the object who has the attraction to the vehicle, while the obstacles are regarded as the objects that generate repulsive force to the vehicle \cite{khatib1985}. All the attractive and repulsive forces are quantified and presented in the form of gravity, where the positive gravity is correlated with the distance between the vehicle position and the destination, and negative gravity is performed within the domain of the obstacles. As the vehicle is closer to the destination, the gravity decreases until it reaches the destination. Deducted by the negative gradient of respective fields, the attractive force $\mathbf{F}_{a}$ and the repulsive force $\mathbf{F}_{r}$ are given by,

\begin{equation}
\mathbf{F}_{a}(x)=-\nabla\mathbf{U}_{a}(x)=k_{a}\rho(x,x_{d})\\   
\label{eq:eq1}
\end{equation}

\begin{equation}
\begin{aligned}
\mathbf{F}_{r}(x)=-\nabla\mathbf{U}_{r}(x)=
\left\{
\begin{array}{lr}
k_{r}(\frac{1}{\rho(x,x_{o})}-\frac{1}{\rho_{o}})\frac{1}{\rho^{2}(x,x_{o})}\nabla\rho(x,x_{o}), \rho(x,x_{d})<\rho_{o}\\
0, \rho(x,x_{d})\geq \rho_{o}
\end{array}
\right.
\label{eq:eq2}
\end{aligned}
\end{equation}

where $-\nabla\mathbf{U}_{a}$ represents the negative gradient of the attractive field; $-\nabla\mathbf{U}_{r}$ represents the negative gradient of the repulsive field; $k_{a}$ is the coefficient for attraction; $\rho(x,x_d)$ represents the distance between the current position $x$ and the destination position $x_d$; $k_r$ is the repulsion coefficient; $\rho(x,x_o)$ represents the distance between the current position to the obstacle position $x_o$ and $\rho_o$ is the radius of the obstacle. 

\begin{figure}[htb]
  \centering
  \includegraphics[width=1.03\textwidth]{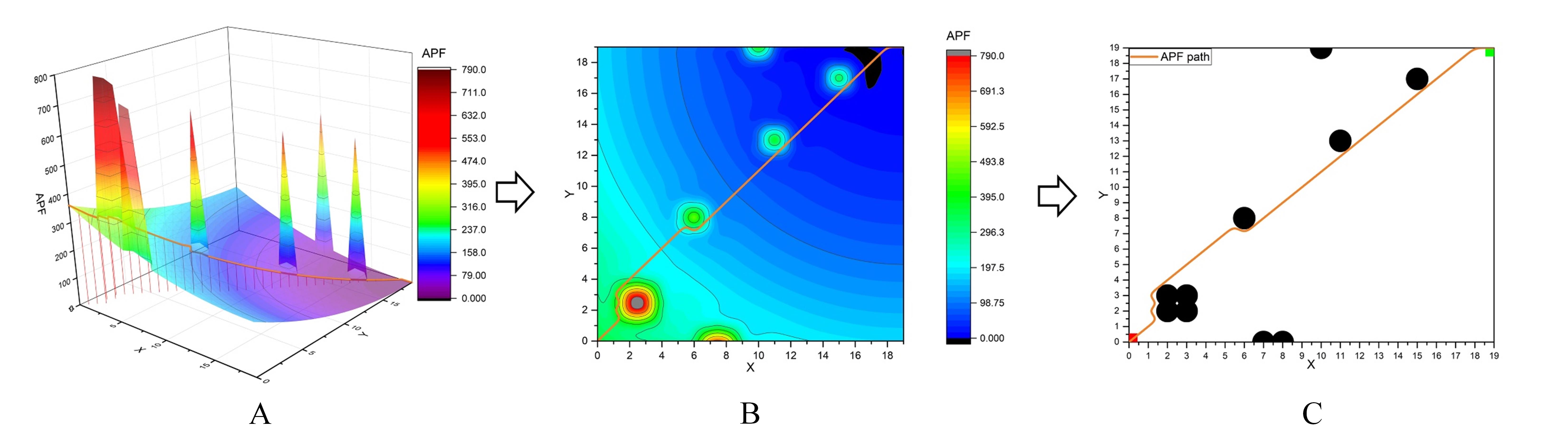}
  \caption{Path derived by the APF method on the 2D modeling map. A: Path planning based on the distribution of the APF on the map, B: APF path on the contour map, C: Final APF path presented on the 2D modeling map \cite{Zhu2021}.}
  \label{fig:fig3}
\end{figure}

Therefore, the destination has the lowest gravity field but the highest gravity force for attraction, while the gravity field for the obstacles performs higher such that the vehicle can flow along the gravity field descending route to complete the optimal path planning, as the path deducted from \autoref{fig:fig3} A to \autoref{fig:fig3} C.

The APF reduces the calculation complexity as well as performs outstanding real-time reactions, which is widely applied in the area of vehicle path planning. The virtual gravitational potential field realizes a fast deduction of the most optimal path to the target without collisions for the vehicle, by following the guidance of resultant forces given by the pre-designed attraction and repulsion \cite{ralli1994}. Zhou and his colleagues improved the APF method with a particle swarm algorithm to increase the path finding efficiency for tangent navigating robots \cite{zhou2018}. Lin and his colleagues designed a subgoal algorithm for the APF such that the path planning of the unmanned vehicle can overcome the local minimum and track the most optimal path \cite{Lin2019}. The decision tree was added to the APF to form the efficient path planning algorithm without local minimum and collisions for vehicles \cite{Lin2020}. Regarding the environmental factors, the effect of ocean currents was then involved in the path planning of the UUV while using the APF method \cite{Zhu2021}. 

However, most of the APF research do not involve the environmental disturbance in the design, thus affecting the practical application of the APF. Moreover, the APF method for vehicle path planning often deduces the problem of local minimum, where the vehicle might stick at halfway instead of reaching the target position due to the larger resultant effect produced by the local minimum point \cite{rahman2017}. The large computation complexity caused by the increasing obstacle numbers also affects the planning efficiency of the APF method.

\paragraph{Intelligent Path Planning Method}
More and more artificial intelligent methods have been applied in the studies of the UUV path planning in recent years, covering the genetic algorithm, swarm intelligence, fuzzy logic and neural network algorithm.

The genetic algorithm (GA) and ant colony algorithm (ACO) were widely used in the early times for the underwater path planning. The GA method imitates the natural selection and evolution procedure to provide the optimal solution through iterations, which has been involved in the path planning and obstacle avoidance under the underwater environment of dynamic currents effect \cite{alvarez2004,Cheng2012}. The ACO method belongs to the swarm intelligence algorithm, where it is designed based on the swarm behavior of ant groups while chasing the food, and the ant behavior-based intelligent method has been proved to work well in the UUV global path planning \cite{RN1346,ant2}. The swarm intelligence methods are broadly applied in the UUV path planning in recent years due to their simple implementation, fast convergence speed and satisfactory robustness while modeling based on different swarming animal groups \cite{Mo2015}. The swarm intelligence algorithms provide outstanding performance in the path planning of the UUV, yet the local minimum problem can be produced by this intelligent method, which finally leads to premature execution before reaching the destination.

The fuzzy logic performs well in the UUV path planning and obstacle avoidance owing to its expertise in processing the information uncertainty while the underwater environment is of high uncertainty and incompleteness \cite{Lee1,Lee2}. Kim and his colleagues used the fuzzy logic-based algorithm to deduct the turning direction and angle of the UUV to avoid collisions and complete the path planning \cite{Kim2006}; Ali developed a fuzzy ontology modeling method to realize the UUV path planning \cite{Ali2015}. The fuzzy logic-based intelligent algorithm need not establish accurate mathematical models as it is derived from the human cognitive experience, so the fuzzy logic can retune itself during the navigation and overcome the local minimum problem. However, the fuzzy logic rule relies heavily on experts' experience and approximations and unverified errors cannot be thoroughly avoided. The complexity of the dynamic environmental factors also challenges the adaptiveness of the fuzzy logic design \cite{leblanc2009}.

\begin{figure}[!t]
  \centering
  \includegraphics[width=0.75\textwidth]{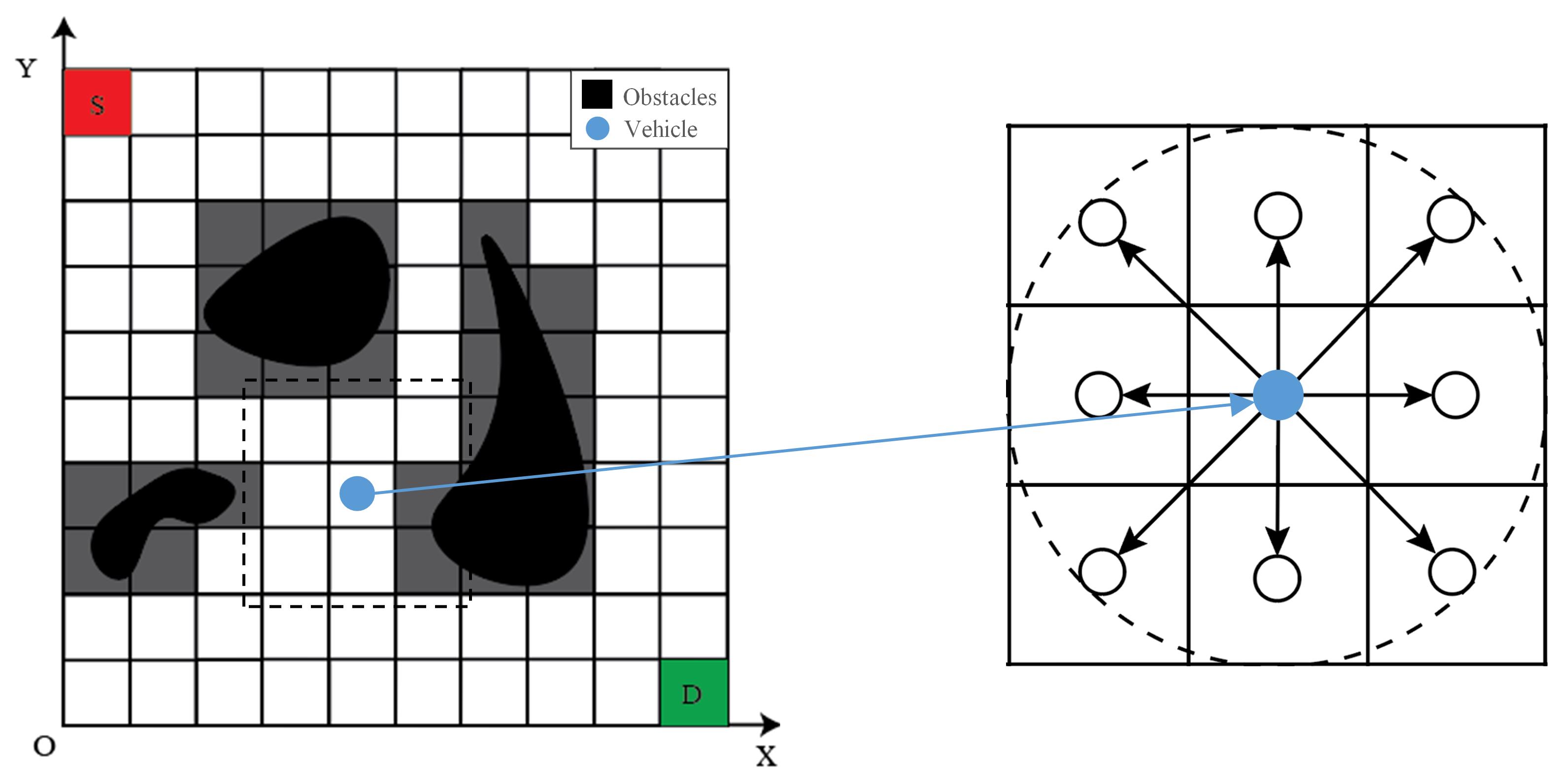}
  \caption{The 2D model of the bio-inspired neural network-based path planning algorithm. S: Start, D: Destination.}
  \label{fig:fig4}
\end{figure}

The application of neural networks in the vehicle path planning has obtained wide attention in recent decades \cite{Sun2017}. Ghatee applied the Hopfield neural network in the optimization of path planning distances \cite{Ghatee2009}. Yang and his group proposed a bio-inspired neural network for the vehicle path planning, where both optimal planning paths and collision avoidance are realized with high efficiency \cite{Yang2009}. The bio-inspired neural network helps to derive the optimal path that composed by continuous coordinates of the vehicle movement, based on a grid-based map and its corresponding neural network model, where each grid represents a single neuron as shown in \autoref{fig:fig4}. The bio-inspired neural network algorithm continuously updates the state of neurons by transmitting the information through the network to give an instant reaction and reduces the complexity by limiting the searching area to a certain range. Therefore, the bio-inspired neural network path planning utilizes the preserved information in the neurons to update its planning design and meanwhile adjust the network on time such that it suits well in the dynamic underwater environment, providing an efficient and high adaptive approach for the UUV path planning \cite{ZhuPP2021}. 

In recent years, the application of reinforcement learning (RL) in the UUV path planning has grown quickly. The RL method updates the vehicle's states and converges to the optimal path planning solution by making actions according to rewards set based on the environment. RL based path planning combined with APF for intervention AUVs has been developed to remove sea urchins at an affordable cost \cite{RLpp22019}. AUV path planning in a complex and changeable environment is achieved through the combination of RL and deep learning \cite{RLpp42021}. Wang and his colleagues proposed a multi-behavior critic RL algorithm for AUV path planning to overcome problems associated with oscillating amplitudes and low learning efficiency in the early stages of training and they reduced the time consumed by the RL algorithm convergence for avoiding obstacles \cite{RLpp52021}. However, the slow convergence issue of RL-based path planning methods still needs further investigation.

To sum up, the methods that are commonly used in the point-to-point path planning of the UUV have been concluded in the Table 2, where their implement theory, benefits and drawbacks are described. Details of various intelligent methods applied on the point-to-point path planning of a UUV can found in the fourth part of Section 2.2.1.

\begin{table}[!t]
  \centering
  \footnotesize
  \caption{Algorithms for UUV point-to-point path planning}
  \label{tab:table2}
  \begin{tabular}{>{\raggedright}p{2.3cm}p{6.2cm}p{2.8cm}p{3.5cm}}
    \toprule
    $\textbf{Algorithms}$ & $\textbf{Logic}$ & $\textbf{Benefits}$ & $\textbf{Drawbacks}$\\
    \midrule
Map Building Method [33-40] & \makecell[l]{Visibility graph-based: \\1) Establish as graph on the connection of the \\vehicle, polygonal obstacle vertex and the desti-\\nation without crossing the obstacles; \\2) Find the optimal path between the origin point \\and the destination point that has the shortest \\distance. \\Grid-based: \\1) Decompose the surrounding area into non-\\overlapping but connected cells; \\2) Address the optimal path between the origin \\and the destination cells without collisions.} & \makecell[l]{1) Easy to implement; \\2) Direct because of \\visible mapping.} & \makecell[l]{Visibility graph-based: \\1) Long time consumption \\when establishing the graph; \\2) Lack flexibility; \\3) Do not work for circular \\obstacles. \\Grid-based: \\1) Large computation; \\2) Lack consideration of \\environmental disturbance.}\\ \addlinespace

Artificial Potential Field [41-47] & \makecell[l]{1) Predefine a virtual artificial potential field; \\2) Assume the destination provides the attractive \\force destination while obstacles generate repul-\\sive force to the vehicle; \\3) Address the optimal path for the vehicle through \\the field descending route.} & \makecell[l]{1) Simple mechanism; \\2) High efficiency and \\real-time reaction.} & \makecell[l]{1) Local minimum; \\2) Sometimes induce large \\computation.}\\ \addlinespace
Intelligent Algorithms(GA, ACO, Fuzzy logic, NN, RL-based) [48-52,55-64] & \makecell[l]{1) Regard the path planning as a search optimi-\\zation problem; \\2) Take the searching cost as the objective func-\\tion; \\3) Optimization through iterations.} & \makecell[l]{1) Easy to implement; \\2) Adaptiveness.} & \makecell[l]{1) Unsatisfactory real-time \\reaction owing to the compu-\\tation complexity; \\2) Local minimum.}\\
   \bottomrule 
  \end{tabular} 
\end{table}

\subsubsection{Full-coverage Path Planning}
The full-coverage path planning has to be considered when the vehicle reaches the designated search area, where the global area of the searching map shall be covered. The goal of the full-coverage path planning for the UUV is to simultaneously realize the high coverage rate, the low repetition route and the short navigating distance. 

The random coverage strategy was proposed at early times to complete the full-coverage path planning. Maxim proposed a full-coverage path planning algorithm for multi-robots in the unknown environment, which need not obtain the global map information in advance and the vehicles would not produce collisions with each other \cite{maxim2002}. However, the random coverage strategy is used in this algorithm to traverse the operating area for each vehicle, hence problems of the path clutter, the high repetition rate and a not complete full-coverage path planning are induced. 

The map building method based on sensor information is combined to achieve a complete full-coverage path planning for the vehicle. Parlaktuna developed a full-coverage path planning method based on the sensor system for multiple vehicles, where the generalized Voronoi diagram was applied for modeling and initializing a full-coverage path and the path section is divided by the capacitated arc routing algorithm \cite{Parlaktuna2009}. The full-coverage path planning is realized by the division of the vehicle navigation area, yet it only suits the map consists of narrow paths such that the vehicle can cover the whole area through one-direction navigation while it does not work well in a large space.

Based on the building map, many researchers have refined the full-coverage path planning method in the multi-vehicle system while resolving the collaboration problem, which is denoted as collaborating full-coverage problem. Janchiv applied cell decomposition to separate the operating area into several sub areas and determine the suitable path planning result for each sub area respectively. The vehicles can consume the least turning times and meanwhile maintain a high efficiency to complete the full-coverage path navigation \cite{Janchiv2011}. However, Janchiv's method did not consider the collaboration among the vehicle groups and the method lacks the proof of robustness. Rekletis introduced the boustrophedon cellular decomposition algorithm into the collaborating full-coverage path planning problem for multiple vehicles, where the domain decomposition method breaks the area and a greedy auction algorithm resolves the task assignment as well as the collaboration of the vehicles \cite{Rekleitis2008}. This path planning method achieves the full coverage of the whole area, yet a large repetition of the navigated paths still cannot be avoided. Hazon proposed the multi-robots spanning-tree coverage algorithm (MSTC) that largely increased the robustness for the multiple vehicles to traverse the whole area while it cannot guarantee the optimal coverage time \cite{Hazon2008}. Therefore, Zheng developed a multi-robots forest coverage (MFC) algorithm that realized the optimal coverage time \cite{Zheng2007}. 

With the advancement of intelligent algorithms, the full-coverage path planning method that can retune or optimize itself has been developed. For instance, Kapanoglu combined genetic algorithm (GA) and template match approach into the collaborating full-coverage path planning problem, where the GA is used to address the best match template for each single vehicle path planning such that both the least traversing paths and optimal coverage time can be promised, but the method lacks the adaption for dynamically changing environment, which is commonly seen for the underwater area of the UUV \cite{Kapano2012}. The advantage of bio-inspired neural network to resolve the collaborating full-coverage path planning problem of ground cleaning robots, where each vehicle regards the others as obstacles such that the full-coverage with collision-free collaboration is realized. However, the large complexity of the neural network is still a big concern \cite{Yang2004,yaocoverage2}. 

Moreover, to increase the efficiency of full-coverage traversing algorithm, studies related to target search algorithms based on probabilistic priority map have been proposed. For example, Cai developed a full-coverage path planning algorithm depending on the bio-innovation such as animal behaviors but considering the probabilistic priority, where the efficiency is increased yet the method is not highly adaptive to the changing environment \cite{Cai2017}. Yao has proposed the full-coverage path planning methods depend on the probability map of targets, where intelligent methods such as biased min-consensus (BN-BMC) algorithm, Gaussian-based analysis or SOM are combined respectively \cite{Yao2021coverage1,yaocoverage3,yaocoverage4}.

Generally, most full-coverage path planning methods are applied on the land or aerial vehicles rather than the UUV. The problems of not complete full coverage and high repetition routes usually occur during the navigation process. The studies on full-coverage path planning for the UUV in underwater environments are concluded in the Table 3, which are still at the very early stage and attention has to be paid to the concerns of enhancing the efficiency of full coverage and decreasing the repetition rate. 

\begin{table}[!t]
  \centering
  \footnotesize
  \caption{Algorithms for UUV full-coverage path planning}
  \label{tab:table3}
  \begin{tabular}{>{\raggedright}p{2cm}p{5.1cm}p{3.5cm}p{3.8cm}}
    \toprule
    $\textbf{Algorithms}$ & $\textbf{Logic}$ & $\textbf{Benefits}$ & $\textbf{Drawbacks}$ \\
    \midrule
Random Coverage Strategy [65] & Traverse the operating area with multiple vehicles following the random coverage strategy. & \makecell[l]{1) No need of initial \\environmental information; \\2) Collision avoidance.} & \makecell[l]{1) Not complete full-coverage; \\2) High repetition.}\\
\addlinespace
Sensor-based map building method [66-69]& \makecell[l]{1) Build the map based on sensor infor-\\mation; \\2) Apply the diagram algorithms for mo-\\deling and initialize a full-coverage path \\by dividing the path into sections \\accordingly.}
& \makecell[l]{1) Complete full-coverage; \\2) Consider multi-vehicle \\collaboration.} & \makecell[l]{1) Only work for narrow paths; \\2) Complete full-coverage \\cannot be realized in conditions \\of broad area; \\3) Lack of robustness; \\4) Lack of optimal multi-vehicle \\task assignment; \\5) High repetition.}\\ \addlinespace
Intelligent method-based full coverage path planning [71-73] & Apply intelligent methods such as GA or NN for each single vehicle path planning. & \makecell[l]{1) Complete full-coverage; \\2) Collision avoidance due \\to self-regulation; \\3) High efficiency of shortest \\covering time and lowest \\energy cost.} & \makecell[l]{1) Low adaptiveness to the \\dynamic environment (GA); \\2) Large computation (NN).}\\ \addlinespace
Probabilistic priority-based full coverage path planning [74-77] & Plan the path due to the predefined probabilistic priority. & \makecell[l]{1) Easy to implement; \\2) Complete full coverage; \\3) Increasing efficiency.} & Not adaptive to dynamic environment.\\
    \bottomrule
  \end{tabular}
\end{table}

\section{Technologies of UUV Tracking Control}
Due to the complex environmental factors of the deep-water space, such as the high pressure, invisibility or unpredictable obstacles, UUVs are applied in most cases when operating undermarine to guarantee the safety and efficiency \cite{r5,r6,TvT2}. Therefore, achieving the robustness and accuracy of controlling the UUV to track the desired trajectory, is dramatically important for completing the real-time underwater navigation \cite{r7,TVT1}. As mentioned in the introduction section, UUVs are mainly divided into ROV and AUV. The ROV can be directly controlled through a control model for propagation, Robot Operating System (ROS) modules, a visual processing pipeline and a dashboard interface for the end-user, where the user gives commands remotely step by step \cite{Ray2021}. This is known as the remote control and the ROV is controlled manually under the case, which is not the critical point of the section as the manual control strategy is direct and simple. For the AUV, the control is realized in an autonomous way, meaning the AUV has to recognize the surrounding areas and make the decision itself. Moreover, some ROVs also support the autonomous mode as the AUV, i.e., the “Falcon" ROV. Hence, in this review, the tracking control technologies emphasize the autonomy of UUVs, and applications on ROVs can also serve as examples of autonomous trajectory tracking control.

To realize the satisfactory trajectory tracking of the UUV, the vehicle must follow the desired path following the corresponding time period. In other words, the errors between the desired and actual trajectories have to be minimized at the different degrees of freedom \cite{r26}. However, different from common unmanned vehicles such as the land vehicle or the unmanned surface vehicle (USV), the UUV system contains more states, whose degrees of freedom (DOF) can be extended to six. 

\begin{figure}[htb]
  \centering
  \includegraphics[width=0.95\textwidth]{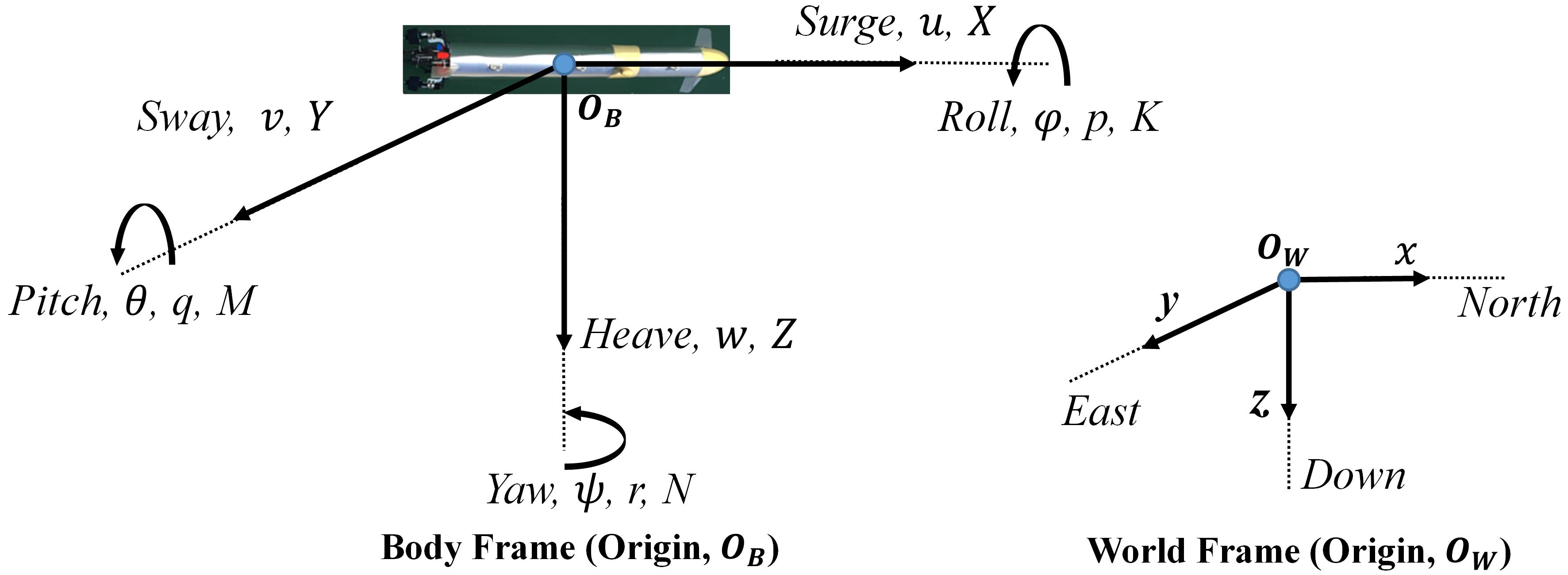}
  \caption{Degrees of freedom and corresponding axes for a UUV.}
  \label{fig:fig5}
\end{figure}

For the kinematic equation of the UUV, the velocity vector $\mathbf{v}$ can be transformed into the time derivative of position vector $\mathbf{\dot{p}}$ by a transformation matrix $\mathbf{J}$ as
\begin{equation}
    \dot{\mathbf{p}}=\mathbf{J}(\mathbf{p})\mathbf{v}.
\end{equation}
where the velocity vector $\mathbf{v}$ is $[u~v~w~r~p~q]^{T}$, as the velocity variable shown at each DOF in \autoref{fig:fig5}.

The transformation matrix $\mathbf{J}(\mathbf{p})$ is
\begin{equation}
    \mathbf{J}(\mathbf{p})=
\begin{bmatrix}
   \mathbf{J}_1 & \mathbf{O}_{3\times 3} \\
   \mathbf{O}_{3\times 3} & \mathbf{J}_2 
  \end{bmatrix},
\end{equation}
where $\mathbf{J}_1$ and $\mathbf{J}_2$ are
\begin{equation}
    \mathbf{J}_1=
\begin{bmatrix}
   \cos\psi\cos\theta & \cos\psi\sin\theta\sin\varphi-\sin\psi\cos\varphi & \cos\psi\sin\theta\cos\varphi+\sin\psi\sin\varphi \\
   \sin\psi\sin\theta & \sin\psi\sin\theta\sin\varphi+\cos\psi\cos\varphi & \sin\psi\sin\theta\cos\varphi+\cos\psi\sin\varphi \\
   -\sin\theta & \cos\theta\sin\varphi & \cos\theta\cos\varphi
  \end{bmatrix},
\end{equation}
\begin{equation}
    \mathbf{J}_2=
\begin{bmatrix}
   1 & \tan\theta\sin\varphi & \cos\varphi\tan\theta \\
   0 & \cos\varphi & -\sin\varphi \\
   0 & \sin\varphi/\cos\theta & \cos\varphi/\cos\theta
  \end{bmatrix}.
\end{equation}
Among the six DOFs of the underwater vehicle, surge, sway, heave, roll, pitch, and yaw, roll and pitch can be neglected since these two DOFs barely have an influence on the underwater vehicle during practical navigation. Therefore, when establishing the trajectory tracking model to keep a controllable operation of the UUV, usually only four DOFs: surge, sway, heave, and yaw are involved (see DOFs shown in \autoref{fig:fig5}). Hence, for the kinematic equation, the position vector can be simplified as 
\begin{equation}
   \mathbf{\dot{p}}= 
 \begin{bmatrix}
   \dot{x} \\
   \dot{y} \\
   \dot{z} \\
   \dot{\psi}
  \end{bmatrix}
=\mathbf{J}(\mathbf{p})\mathbf{v}=
\begin{bmatrix}
   \cos{\psi}& -\sin{\psi} & 0 & 0 \\
   \sin{\psi}& \cos{\psi} & 0 & 0 \\
   0 & 0 & 1 & 0 \\
   0 & 0 & 0 & 1
  \end{bmatrix}\mathbf{v}
  =\begin{bmatrix}
   \cos{\psi}& -\sin{\psi} & 0 & 0 \\
   \sin{\psi}& \cos{\psi} & 0 & 0 \\
   0 & 0 & 1 & 0 \\
   0 & 0 & 0 & 1
  \end{bmatrix}
  \begin{bmatrix}
   u \\
   v \\
   w \\
   r
  \end{bmatrix}, 
\end{equation}
where $\mathbf{J}$ is a transformation matrix derived from the physical structure of the UUV body, $[u~v~w~r]^T$ represents the velocities at the chosen four axes of a UUV, as presented in \autoref{fig:fig5}.

For UUV, a generally accepted dynamic model has be defined as
\begin{equation}
    \mathbf{\dot{v}}+\mathbf{C}(\mathbf{v})\mathbf{v}+\mathbf{D}(\mathbf{v})\mathbf{v}+\mathbf{g}(\mathbf{p})=\boldsymbol{\tau}\,,
\end{equation}
where $\mathbf{M}$ is the inertia matrix of the summation of rigid body  and added mass; $\mathbf{C}(\mathbf{v})$ is the Coriolis and centripetal matrix of the summation of rigid body and added mass; $\mathbf{D}(\mathbf{v})$ is the quadratic and linear drag matrix; $\mathbf{g}(\mathbf{p})$ is the matrix of gravity and buoyancy; and $\boldsymbol{\tau}$ is the torque vector of the thruster inputs. 

The torque vector of the thruster input is represented by
\begin{equation}
    \boldsymbol{\tau}=
    \begin{bmatrix}
       \tau_x & \tau_y &
       \tau_z & \tau_k & \tau_m &
       \tau_n
    \end{bmatrix}^T,
\end{equation}
where $\tau_x$, $\tau_y$, $\tau_z$, $\tau_k$, $\tau_m$ and $\tau_n$ representing torques of the UUV at surge, sway, heave, pitch, roll and yaw directions, as shown in \autoref{fig:fig5}. In addition, as is mentioned in the previous section, in some practical cases, the torques at pitch and directions can be neglected. 

Due to the nonlinearity of the UUV system, the typical linearization method, proportion integration differentiation (PID) control does not work very well and is less studied for the UUV trajectory tracking \cite{Li2022}. Hence in this section, the major methods that are used for the UUV trajectory tracking are concluded and categorized into the conventional control (consists of the backstepping control, sliding mode control and model predictive control), intelligent control and the fault-tolerant control.

\subsection{Conventional Control}
In this section, some conventional control such as the backstepping control, sliding mode control and model predictive control are described. Studies regarding their applications in the trajectory tracking control of the UUV are stated. The conclusion of the features for these conventional controls is given in Table 4.

\subsubsection{Backstepping Control}

In the backstepping method, control functions for each subsystem are designed based on the Lyapunov techniques and generated to form the complete control law \cite{r28}.\\\\
However, the actuator saturation is induced by the speed-jump problem, which usually occurs in the backstepping control methods for trajectory tracking \cite{r8}. The excessive speed references affect the robustness of the UUV trajectory tracking by introducing excessive fluctuations of velocities at initial states or other large error states during the kinematic controlling procedure. Therefore a sharp speed change is derived from the large errors accumulated from the generation of the subsystems, where speed-jump issues are induced when the deviation occurs. As the UUV cannot provide infinite driving inputs such as torques/forces due to its underwater workspace and limited electric power, the actuator saturation, has to be considered during the trajectory tracking process of the vehicle, with the torques/forces constraints applied\cite{r9, r10, ry2}.

\subsubsection{Sliding Mode Control}
As one of the most basic robust controlling strategies, the sliding mode control (SMC) is widely used due to its simple and robust mechanism, hence the SMC is often chosen to construct the trajectory tracking controller of the vehicle \cite{rcn1,rcn2}. In the SMC, a sliding surface mode is supposed to follow the desired tracking and keep the controlled outputs remaining on the surface. Once the controlled trajectory is out of the perfect sliding surface mode, the SMC will push the trajectory slide back to the surface with addition or subtraction on the original controlling equation \cite{r17,r18}. Therefore the SMC restricts the fluctuation of controlled outputs in an acceptable range through a simple operation, which is highly applicable in the trajectory tracking problems\cite{r19}. 

However, the SMC suffers chattering issues though it performs robust to variable changes, which is a critical factor that needs to be considered when designing the control strategy \cite{Slotine1986}. Xu refined the SMC with a bio-inspired neural network algorithm such that the chattering problem can be alleviated, yet it is limited to the application of land vehicles where fewer degrees of freedom are involved \cite{Xu2021}.

\subsubsection{Model Predictive Control}
Model predictive control (MPC) is appropriate for the UUV system that navigates in the mode of slow velocity. MPC is not demanding on the model accuracy and meanwhile provides in-time feedback, and constraints can be added to the control strategy to alleviate the jumps of the speed. Therefore, motivated by the requirements of in-time reaction and restriction of velocities within physical constraints throughout the whole tracking process, the MPC control stands out to be one of the most feasible solutions for constructing the tracking control for the UUV \cite{mpvreview2}. 

The MPC resolves the online optimization problem at each timeslot and derives in-time predictions with minimum errors \cite{mpc1}. The optimization process performs a receding horizon in the MPC. When deducting the solution of the next timeslot, the optimization algorithm embedded in the control system first gives an optimized sequence within a pre-defined timeslot. The first result of the sequence is adopted as the solution and works as the basis for the next optimization loop while time receding \cite{mpcreview1}. At the same time, constraints are added in the optimization to set the limitation to the optimized results as well as the variation of the sequence results \cite{mpc1,mpc2}. By this receding optimization algorithm and the set constraints, online control can be realized and excessive velocity results are avoided. Bing and his colleagues applied MPC as the vehicle trajectory tracking control and satisfactory tracking results of fewer and gentle fluctuations are achieved, which demonstrates the effectiveness of the MPC \cite{r8}.  
\subsection{Intelligent Control}
Intelligent controls refer to the control strategies that can realize desired control goals without manual interventions, which are often used under situations of large uncertainties. 

The fuzzy logic system is used as a component of the intelligent control, which addresses the uncertainties and gives a more flexible criterion for obtaining the optimized predictions within its conceptual framework \cite{Jing,Fang}. It can also limit the output data and smoothen the kinematic error curves derived from the conventional backstepping method through its decision function. Compared to the MPC, the fuzzy logic controller constructs a model that imitates the human decision-making with inputs of continuous values between 0 and 1, which largely simplifies the computing process \cite{Lee1,Lee2}. Some researchers have achieved successful tracking based on the fuzzy logic-refined backstepping method, yet their application is based on the underactuated surface vehicle (USV), with fewer states involved compared to the UUV \cite{Ning}. Some researchers have applied synergetic learning in their controller designed for vehicles and better performance is obtained, but they do not consider practical constraints of the vehicle \cite{tdcs_synergetic}. Li has developed the fuzzy logic-based controller that provides satisfactory tracking results even with time-varying delays or input saturation, but the effectiveness of the algorithm on specific models such as the UUV has not been discussed \cite{Yongming}. Wang and his colleagues developed a fuzzy logic-based backstepping method, yet it has not been experimented under specific application scenarios, with dynamic constraints applied \cite{r22}.

\begin{figure}[!t]
  \centering
  \includegraphics[width=0.95\textwidth]{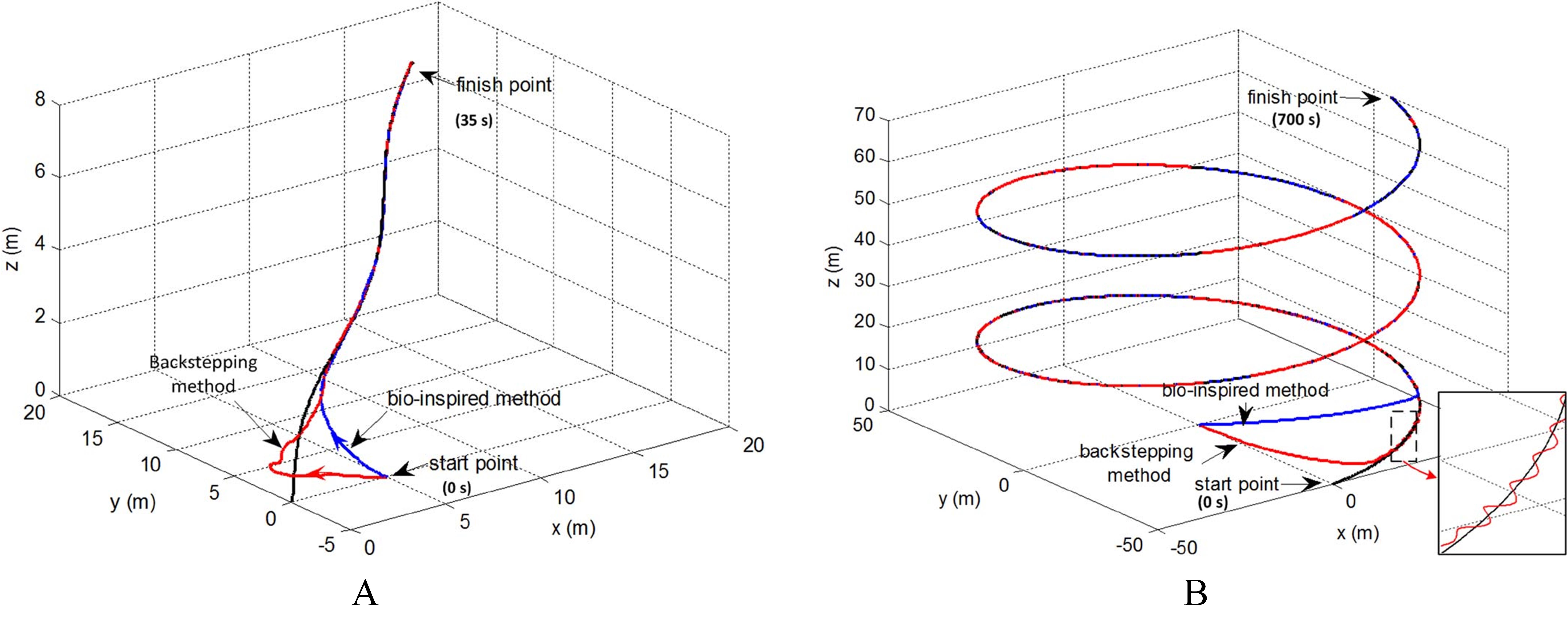}
  \caption{Tracking trajectory comparison of the bio-inspired model based control and conventional backstepping control for the underwater robots. A: curve tracking, B: helix tracking \cite{Sun_2014_7000m}.}
  \label{fig:fig6}
\end{figure}

As a typical intelligent method, the neural network-based models have been applied to the tracking control of the UUV for many years \cite{r5}. Due to the complex underwater work environment and limited electric power of UUVs, the excessive speed references as well as the actuator saturation problems have to be considered. The bio-inspired backstepping controller was introduced in the control design to give the resolution respectively \cite{r10}. Based on the characteristics of the shunting model, the outputs of the control are bounded in a limited range with a smooth variation \cite{Zhu_2013sliding}. The bio-inspired backstepping controller has been applied on different UUVs under various conditions by combining with a sliding mode control that controls the dynamic component of the vehicle. An adaptive term is used in the sliding mode control to estimate the nonlinear uncertainties part and the disturbance of the underwater vehicle dynamics \cite{Sun_2012sliding}. For example, the actuator saturation problem of a 7000m manned submarine was resolved through this bio-inspired backstepping with the sliding mode cascade control \cite{Sun_2014_7000m}. The control contains a kinematic controller that uses a bio-inspired backstepping control to eliminate the excessive speed references when the tracking error occurs at the initial state. Then, a sliding mode dynamic controller was proposed to reduce the lumped uncertainty in the dynamics of the UUV, thus realizing the adaptive trajectory tracking control without excessive speeds for the vehicle, as the satisfactory curve and helix tracking results shown in \autoref{fig:fig6}. Jiang accomplished the trajectory tracking of the autonomous vehicle in marine environments with a similar bio-inspired backstepping controller and the adaptive integral sliding mode controller \cite{Jiang_2018ocean}. In the sliding mode controller, the chattering problem was alleviated, which increased the practical feasibility of the vehicle. However, more studies are needed to compare to prove the effectiveness of the proposed control strategy, such as the tracking control based on the filtered backstepping method. 

GA methods are also applied in the intelligent controls of UUVs. They are usually applied based on the aim of addressing the most optimal solution during the control process owing to their feature of self-evolution. However, the computation cost of the GA methods always adds a burden to the tracking control algorithms such that they are usually combined with other intelligent algorithms to reach a more efficient control strategy. Tavanaei-Sereshki applied quantum genetic algorithm (QGA), an optimization algorithm based on the probability that combines the idea of quantum computing and traditional genetic algorithm to realize the UUV's tracking along desired paths \cite{controlGA32018}. Zhang described a route planner that enables an AUV to selectively complete part of the predetermined tasks in the operating ocean area when the local path cost is stochastic through a greedy strategy based GA (GGA), which includes a novel rebirth operator that maps infeasible individuals into the feasible solution space during evolution to improve the efficiency of the optimization and use a differential evolution planner for providing the deterministic local path cost \cite{controlGA12022}. 

The brief conclusion of intelligent controls on UUVs can be found in Table 4, details of various intelligent methods on tracking control of UUV are described in Section 3.2.

\begin{table}[!t]
  \centering
  \footnotesize
  \caption{Algorithms for UUV trajectory tracking}
  \label{tab:table4}
  \begin{tabular}{>{\raggedright}p{2cm}p{6.3cm}p{3.2cm}p{3.2cm}}
    \toprule
    $\textbf{Algorithms}$ & $\textbf{Logic}$ & $\textbf{Benefits}$ & $\textbf{Drawbacks}$ \\
    \midrule
Backstepping Control [84-88] & Design control function by generating the subsystem established based on the Lyapunouv therom. & \makecell[l]{1) Easy to implement for \\nonlinear system; \\2) Real-time reaction.} & \makecell[l]{1) Unsatisfactory adaptive-\\ness; \\2) Produce excessive speed \\references and actuator sa-\\turation problems.}\\ \addlinespace

Sliding Mode Control [89-95] & \makecell[l]{1) Suppose a sliding surface mode to follow the \\desired tracking; \\2) Keep the controlled outputs remain on the \\surface.} & \makecell[l]{1) Simple algorithm; \\2) Robust.} & Chattering issue. \\ \addlinespace

Model Predictive Control [96-99,112] & \makecell[l]{1) Resolve the online optimization problem in \\each timeslot and derive in-time predictions with \\minimum errors; \\2) Optimization algorithm embedded in the control \\system gives an optimized sequence within the \\pre-defined timeslot; \\3) The first result of the sequence is adopted as \\the solution and worked as the basis for the next \\optimization loop while time receding.} & \makecell[l]{1) No need of high \\system accuracy; \\2) In-time reaction; \\3) Adaptive.} & Long time consumption due to the recursive computation.\\
\addlinespace

Intelligent Control (Fuzzy logic, NN, GA-based) [100-111] & \makecell[l]{1) Embed intelligent algorithms as a search \\optimization for the desired tracking result; \\2) Take searching cost as the objective function; \\3) Optimization through iterations.} & \makecell[l]{1) Easy to implement; \\2) Adaptive.} & Large computation.\\
    \bottomrule
  \end{tabular}
\end{table}

\subsection{Fault-tolerant Control}
\begin{figure}[htb]
  \centering
  \includegraphics[width=0.65\textwidth]{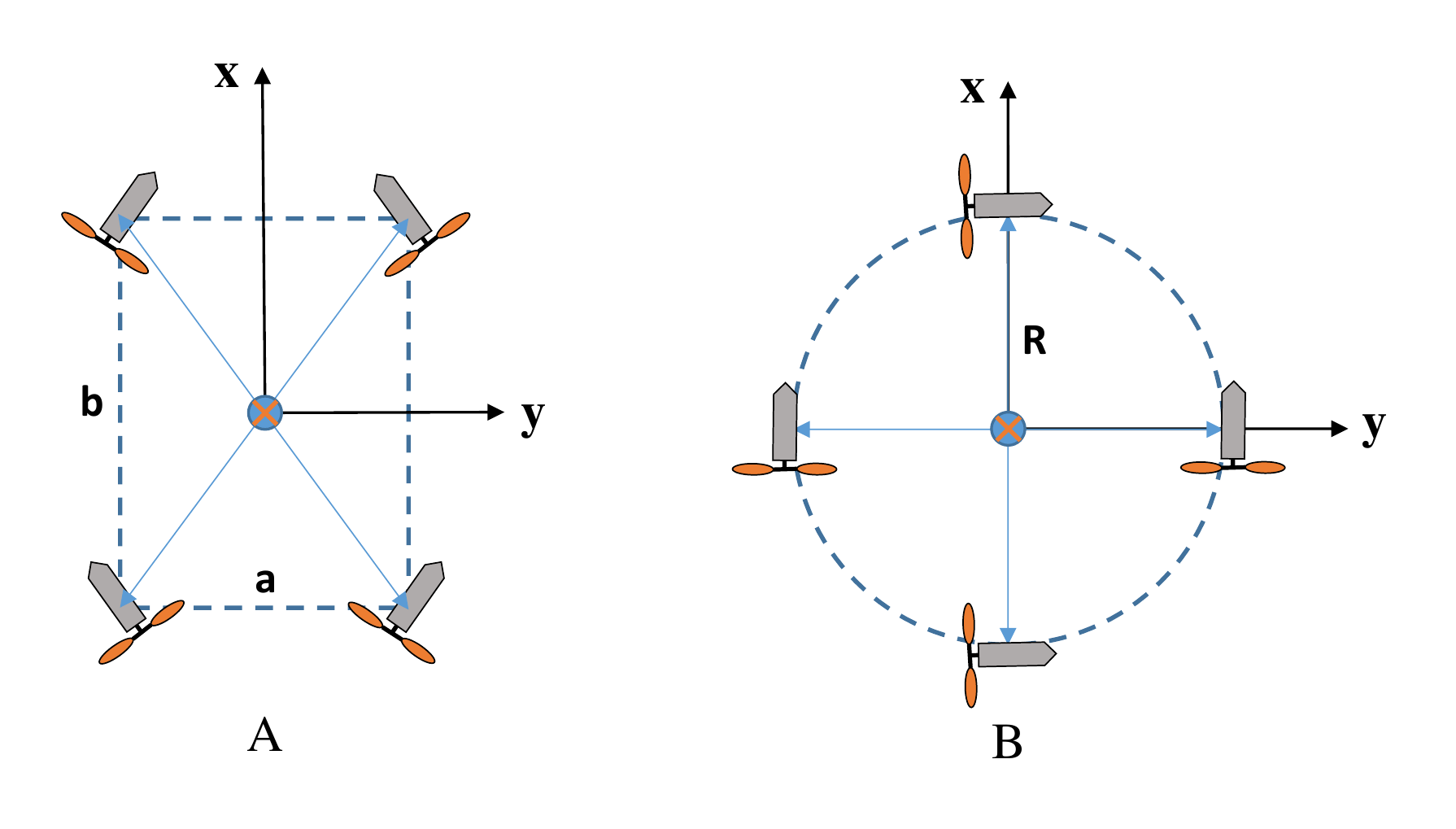}
  \caption{Two typical thruster configurations for the UUV. A: Falcon UUV; B: URIS UUV.}
  \label{fig:fig7}
\end{figure}

Regarding the unpredictability of the underwater environment, it is of high possibility for the UUV to meet unexpected accidents that affect the preset model of the vehicle. For example, in some cases one or more of the UUV's thrusters are out of order, the model has to be modified respectively to continue the desired trajectory tracking designed as before. the fault-tolerant control (FTC) is usually applied to alleviate abrupt errors and provides the most feasible solution when inevitable damages happen to the equipment in different fields \cite{Cao2021}. However, the research regarding the FTC on underwater vehicles has not been thoroughly investigated due to the complexity brought by the underwater environment and the UUV system \cite{Seto2020,ocftc1,TITS1_MSV2021}. \\\\
Several techniques on the FTC were developed in this century \cite{xin2014,TSMC1,Lu2016,Meyer2018}. Based on these studies, the design of the excessive number of thrusters compared to the number of degrees of freedom (DOF) is raised and accepted as a resolution to the UUV FTC problem, which is called the thruster control matrix reconfiguration \cite{martynova2020,ocftc2}. For example, as shown in \autoref{fig:fig7}, the Falcon and URIS UUV have five thrusters while only four DOFs are considered such that the reconfiguration method can be applied. For example, when unexpected fault cases of the vehicle thrusters occur, the thrusters installed on the vehicle that exceeds the number of DOFs (six: surge, sway, heave, row, pitch and yaw) have enough flexible space to be retuned to provide the required propulsion at corresponding DOFs. To implement the thruster control matrix configuration theory in practical cases, the weighted pseudo-inverse matrix method was proposed, where the fault cases are quantified as degrees of damage and serve as the inputs to form the thruster control matrix configuration model \cite{Omerdic2004}. By this method, the process of the FTC is largely simplified, as the required thruster propulsion can be deducted directly through a weighted pseudo-inverse matrix model. Nevertheless, physical constraints of the thruster outputs are rarely considered, thus inducing the over-actuated vehicle issue \cite{Podder2000, ZhangFTC2021}. Additionally, among these studies, most of them work on eliminating the static errors induced by the fault cases. While in UUV practical application, the realization of dynamic control on the vehicle's outputs in a real-time manner, which commonly refers to the trajectory tracking control for underwater vehicles, is of crucial importance \cite{Shen2018,TSMC2}. 

\section{Challenges and Prospects of UUV Motion Planning and Tracking Control}
The motion planning and tracking control of the UUV has a promising future in the maritime projects of underwater rescue, detection, investigation, tube pavement, creature study and military strategy. Hence, there is still a large requirement for thorough and systematic research into the field. In this section, the challenges and prospects of the UUV motion planning and tracking control for the underwater navigation are listed and analyzed.
\subsection{Multi-UUV Collaboration}
In this section, possible future studies on the multi-UUV collaboration are given, mainly divided into the multi-UUV collaboration considering the environmental effect and heterogeneous vehicles, and the hunting of dynamic targets.

\subsubsection{Environmental Effect and Heterogeneous Vehicles}
Most researchers discuss the UUV assignment in an ideal underwater environment and regard the vehicle as a pure particle, which lacks the consideration of the practical condition of the UUV operation. Therefore, the complex environmental factors such as ocean currents effect, unpredictable seamounts and various moving obstacles shall be involved in further studies. Moreover, the UUV system contains uncertainty that cannot be addressed initially, such that the heterogeneous vehicles of different model parameters, navigating velocities or safe distance are required to be studied. To step further, the topics on the formation control of heterogeneous vehicles between the UUV and the USV (or even unmanned aerial vehicles) have become more attractive, as the USV can help to instruct the UUV in real-time positioning through its less-affected GPS system and efficient communication above the water surface \cite{Stewart2018,QYang2020}.

\subsubsection{Dynamic Targets}
Most current UUV motion planning and tracking control studies concentrate on tackling static targets, such as the search of underwater crash transports, yet the following or hunting of dynamic targets for the UUVs is a crucial issue in the military defense for the marine system \cite{Liang2020}. The following or hunting of the dynamic targets covers the topics on the dynamic task assignment, the intelligence of the moving target, the path planning while chasing the target and the containment of the target completed by the UUVs, which are still waited to be further investigated.
 
\subsection{Efficient Underwater Positioning and Path Planning}
The developed underwater positioning methods of real-time efficiency are required, as the conventional positioning and navigating systems like GPS are not valid due to the poor communication affected by the underwater environment. In addition, path planning in the underwater condition is the most vital part of the UUV motion planning, where some innovative intelligent planning methods can be applied in this field to improve efficiency, such as reinforcement learning.

\subsection{Robust Underwater Trajectory Tracking}
Owing to the complexity and uncertainty of the underwater environment and the UUV nonlinear system, the robustness of the UUV trajectory tracking has to be advanced to guarantee the UUV navigates as desired. Therefore, the problems that affect the UUV trajectory tracking robustness such as excessive speed references, actuator saturation and thruster damages are worth investigating \cite{Chao2017}.

\subsection{Real-time Underwater Recognition}
Current approaches that can attain the identity information of the underwater targets are limited due to the low invisibility and unpredictable obstacles of the surrounding environment; the inefficient communication through the fluids in the deep-sea area also restricts the real-time recognition of the underwater target, thus affecting the real-time UUV navigation to unknown targets. Hence the advanced underwater target detection techniques such as the multi-sensor information infusion approach and deep learning-based image recognition can be applied in this field to process and achieve the results in time \cite{Reed2006,Coiras2007,Williams2015,Hou2016}.

\section{Conclusion}

The paper has discussed the methodologies that can be applied to perform a satisfactory UUV motion planning and tracking control, as well as the authors' thoughts on the benefits or drawbacks of these methods. In general, the motion planning and tracking control for UUVs require the vehicle to realize an efficient and robust underwater operation of addressing and approaching the targets, with the optimal planned paths, task assignment among multiple vehicles and robust trajectory tracking procedure. The framework and current investigations of UUV motion planning and tracking control are given in the second and third sections. Moreover, although researchers have developed some effective methodologies on these topics, challenges remain to be resolved, which are listed in the fourth section, together with the possible developments of the UUV motion planning and tracking control technologies.

\bibliographystyle{unsrt}  
\bibliography{references}  






\end{document}